\documentclass{article}

\usepackage{listings}
\usepackage{xcolor}

\usepackage[final]{neurips_2022}
\usepackage{color}
\usepackage{times}
\usepackage{epsfig}
\usepackage{graphicx}
\usepackage{amsmath}
\usepackage{amssymb}
\usepackage{amsmath}
\usepackage{amssymb}
\usepackage{booktabs}
\usepackage{graphicx}
\usepackage{fixfoot}
\usepackage[ruled]{algorithm}
\usepackage[noend]{algpseudocode}
\usepackage{amsfonts}
\usepackage[mathscr]{eucal}
\usepackage{multirow}
\usepackage{comment}

\usepackage{xcolor} 

\usepackage[utf8]{inputenc} 
\usepackage[T1]{fontenc}    
\usepackage{hyperref}       
\usepackage{url}            
\usepackage{booktabs}       
\usepackage{amsfonts}       
\usepackage{nicefrac}       
\usepackage{microtype}      
\usepackage{xcolor}         
\usepackage{graphicx}

\lstdefinestyle{light}{
    language=Python,
    backgroundcolor=\color{white},   
    basicstyle=\ttfamily\small,      
    keywordstyle=\color{blue}\bfseries,    
    commentstyle=\color{gray}\itshape,      
    stringstyle=\color{orange},              
    numberstyle=\tiny\color{gray},           
    numbers=left,                             
    stepnumber=1,                             
    numbersep=5pt,                            
    frame=single,                             
    rulecolor=\color{black},                  
    tabsize=4,                                
    captionpos=b,                             
    breaklines=true,                          
    breakatwhitespace=false,                  
    showspaces=false,                         
    showstringspaces=false,                   
    showtabs=false,                           
    morekeywords={*,...}                      
}

\newif\ifincludefigures

\def\include_figure{1} 

\ifnum\include_figure=1
    \includefigurestrue
\else
    \includefiguresfalse
\fi

\ifincludefigures
\else
    \excludecomment{figure}
\fi

\title{
Language Models and Cycle Consistency for Self-Reflective Machine Translation
}

\author{%
Jianqiao Wangni\\
Tsinghua University \\
Beijing, China\\
  \texttt{zjnqha@gmail.com} 
}

\begin{document}

\maketitle

\begin{abstract}
This paper introduces a novel framework that leverages large language models (LLMs) for machine translation (MT). We start with one conjecture: \textit{an ideal translation should contain complete and accurate information for a strong enough LLM to recover the original sentence.} We generate multiple translation candidates from a source language \( A \) to a target language \( B \), and subsequently translate these candidates back to the original language \( A \). By evaluating the \textit{cycle consistency} between the original and back-translated sentences using metrics such as token-level precision and accuracy, we implicitly estimate the translation quality in language \( B \), without knowing its ground-truth. This also helps to evaluate the LLM translation capability, only with monolingual corpora. For each source sentence, we identify the translation candidate with optimal cycle consistency with the original sentence as the final answer. Our experiments demonstrate that larger LLMs, or the same LLM with more forward passes during inference, exhibit increased cycle consistency, aligning with the LLM model size scaling law [\cite{kaplan2020scaling}] and test-time computation scaling law [\cite{snell2024scaling}]. This work provide methods for, 1) to implicitly evaluate translation quality of a sentence in the target language, 2), to evaluate capability of LLM for any-to-any-language translation, and 3), how to generate a better translation for a specific LLM.   
\end{abstract}

\section{Introduction}
Machine Translation (MT) has been a cornerstone of natural language processing, facilitating globalization by cross-linguistic communication and democratizing newest information access to all population. In recent years, transformer-based large language models (LLMs) have fundamentally changed the field of natural language processing. Introduced by [\cite{vaswani2017attention}], the transformer architecture facilitates parallel processing of input word tokens, significantly improving computational efficiency and successfully scales to unseen model size. Strong language capabilities beyond human imagination emerges from LLM scaling, and create an image of Silicon intelligence for the first time. Transformer-based LLMs can be categorized into several paradigms: encoder-only architectures, like BERT [\cite{devlin2018bert}], which focus on meaningful embeddings of input sequences, and don't fit translation task; the rest two architectures, the encoder-decoder architectures, like T5 [\cite{raffel2020exploring}], which separately process input and output sequences, were born for translation; and decoder-only architectures, like GPT [\cite{radford2019language}], which generate text in an autoregressive manner. Although GPT is not initially trained specifically for translation tasks like T5, it is exceptionally well-suited for a wide range of natural language processing (NLP) tasks via supervised fine-tuning on downstream tasks, including translation, and provides applications to users via prompting [\cite{brown2020language}].

On a related subject, MT evaluation poses significant challenges as there is no unit test for human languages, like Python or Java. Machine-based evaluation metrics, such as BLEU [\cite{papineni2002bleu}], METEOR [\cite{banerjee2005meteor}], and TER [\cite{snover2006study}], provide quantitative assessments based on N-gram overlaps and edit distances. While these metrics offer consistency and objectivity, they often fail to fully capture the semantic adequacy and fluency of translations [\cite{liu2022pre}], although not being a problem for LLMs as they in most cases generate fluent sentences. Higher-level criteria beyond these metrics, such as overall quality and asceticity, nuanced subtleties, professional terminologies and informal idioms, require evaluation from native speaker of the target language, which is expensive and impractical during online inference. Additionally, evaluation on low-resource languages [\cite{luong2015effective}] poses further difficulties due to the data scarcity.

The starting point of our work is a simple conjecture: a good translation,  and the LLM who translated it, should be able to jointly recover the original sentence completely and accurately. This is natural from information theory view of point. For instance, if an English sentence is translated into French and then back to English, a high degree of similarity between the original and final English sentences indicates a more accurate and reliable translation. To prove this, we propose translation \textit{cycle consistency} as a meaningful metric to evaluate translation quality without parallel corpora in source language \( A \) and target language \( B \). By translating a sentence from language \( A \) to language \( B \) and then back to \( A \), we can quantitatively measure the alignment, similarity, or closeness, between the original and back-translated texts.  This method not only economically scale MT assessments, it also streamlines the evaluation process, making it widely usable for offline or online evaluation.

Cycle consistency brings chance for further improving MT through a self-reflective mechanism: to think ahead a few steps. If the evaluation of translation during inference is accurate enough, we can afford to generate multiple candidates and select the best one. Unlike LLM decoding techniques such as beam search, which select the most probable translation based on a search space over a few tokens, solely based the LLM itself, cycle consistency allows for a complete evaluation of all translated tokens, selecting the most coherent and accurate output, with a math formula-backed metric. This is analogous to AlphaGo, which simulates several future steps to determine the best possible action in the current move [\cite{silver2016mastering}]. In this paper, 
we formalize our idea, and empirically investigate the scaling effects, observing that larger models exhibit improved cycle consistency in translations [\cite{chen2021evaluating}], which in turn proves that cycle consistency is a valid and novel metric for evaluation.
\section{Methodology}
Pretrained LLMs like GPT are trained to perform machine translation via prompting. Let \( S_A \) be the original sentence in language \( A \) that we want to translate. The input can be represented as:
\[
S_A = \{ w_1, w_2, \ldots, w_l \}
\]
where \( w \) represents the tokens, and \( l \) is the total number of tokens. To perform machine translation, we construct a prompt \( P \) with a prefix that guides the GPT model to produce a translation in language \( B \). The prompt can be defined as:
\[
P = \text{``Translate the following sentence from language } A \text{ to language } B\text{: ''} + S_A
\]

The GPT model takes the prompt \( P \) as input and generates a sequence of tokens in language \( B \), denoted as \( S_B \):
\[
S_B = \{ w'_1, w'_2, \ldots, w'_m \} = \text{Translate}(S_A; A \rightarrow B; \theta)
\]
where \( w' \) represents the tokens of the translated sentence in language \( B \), \( m \) is the total number of tokens generated, and $\theta$ is all of the decoding hyper-parameters of LLM, including temperature, number of beams (beam-search), top-K candidates, and even the random seed.

\subsection{Self-Reflective Translation}

The self-reflective framework involves thinking ahead, which includes a two-step translation process. The {Forward Translation} is to translate the original sentence \( S_A \) from language \( A \) to language \( B \), generating multiple translation candidates \( \{ S_{B}^{(i)} \}_{i=1}^N \):
    \[
    \{ S_{B}^{(i)} \}_{i=1}^N = \{\text{Translate}(S_A; A \rightarrow B; \theta_i) \}_{i=1}^N.
    \]
The key point in this step is to create enough diversity in answers while not sacrificing accuracy. With many hyper-parameters such as sampling temperatures are unknown, there is also a good chance of iterating over these hyper-parameters within the forward budget. The {Backward Translation} is to translate each hypothesis \( S_{B}^{(i)} \) back to language \( A \), resulting in back-translated sentences \( \{ S_{A}^{(i)} \}_{i=1}^N \):
    \[
    S_{A}^{(i)} = \text{Translate}(S_{B}^{(i)}; B \rightarrow A; \theta), \quad \forall i \in \{1, \dots, N\}
    \]
The consistency \( C^{(i)} \) between the original sentence \( S_A \) and each back-translated sentence \( S_{A}^{(i)} \) is then measured using evaluation metrics such as BLEU score, precision, and accuracy:
\[
C^{(i)} = M(S_A, S_{A}^{(i)})
\]
The translation hypothesis \( S_{B}^{(i^*)} \) corresponding to the highest consistency score is selected as the final translation:
\[
i^* = \arg\max_{i} C^{(i)} \quad \text{and} \quad S_{B}^{(i^*)} \rightarrow \text{Selected Translation}.
\]

\subsection{Metrics for Consistency}

The consistency between two sentences were measured using different standards. We can focus on the precision, the proportion of correctly translated words [\cite{pradeep2018linguistic}]; or focus on the accuracy, the exact match rate between the original and back-translated sentences [\cite{brown2020language}]; we can use BLEU (Bilingual Evaluation Understudy) Score [\cite{papineni2002bleu}], which is the  N-gram overlap:

\[
\text{BLEU} = \lambda_{BP} \times \exp\left(\sum_{n=1}^{N} w_n \log p_n\right)
\]
where \( \lambda_{BP} \) is the brevity penalty for shorter candidate translations, \( p_n \) is the modified N-gram precision for N-grams of order \( n \) in the candidate translation, and \( w_n \) are the weights for each precision score, typically set to \( {1}/{N} \) for equal weighting of N-grams from 1 to \( N \).

The BLEU score ranges from 0 to 1, with a higher score indicating better quality translations that more closely resemble the reference texts. However, while BLEU is effective for capturing N-gram overlap, it has limitations in evaluating semantic meaning and context, which can sometimes lead to misleading results in translation quality assessments [\cite{papineni2002bleu}.

Although there is no golden standard of the consistency, in our experiments, we mainly use ROUGE (Recall-Oriented Understudy for Gisting Evaluation) [\cite{lin2004rouge}, which complements BLEU by focusing on recall-based metrics, evaluating the overlap of N-grams and longest common subsequences between the candidate and reference. 


\subsection{Foundation Model Architecture}

In the landscape of machine translation, GPT [\cite{radford2019language} and T5 [\cite{raffel2020exploring} represent two distinct transformer-based architectures and training paradigms for machine translation. GPT learns a much wider range of knowledge before focusing on translation tasks, often leading to a higher level of human-like intelligence in processing language, albeit at the cost of a larger parameter size. In contrast, T5 utilizes a full encoder-decoder transformer architecture, inherently suitable for machine translation under small model size, and is capable of the task after training on smaller datasets compared to GPT. GPT, on the bright side, has higher potential for difficult translations, given enough pretraining data and model size. Their distinct strengths encourage us to combine both architectures with chain-of-thought reasoning in forward translation.

\section{Experiments}
\begin{figure}[ht]
    \centering
    \includegraphics[width=\linewidth]{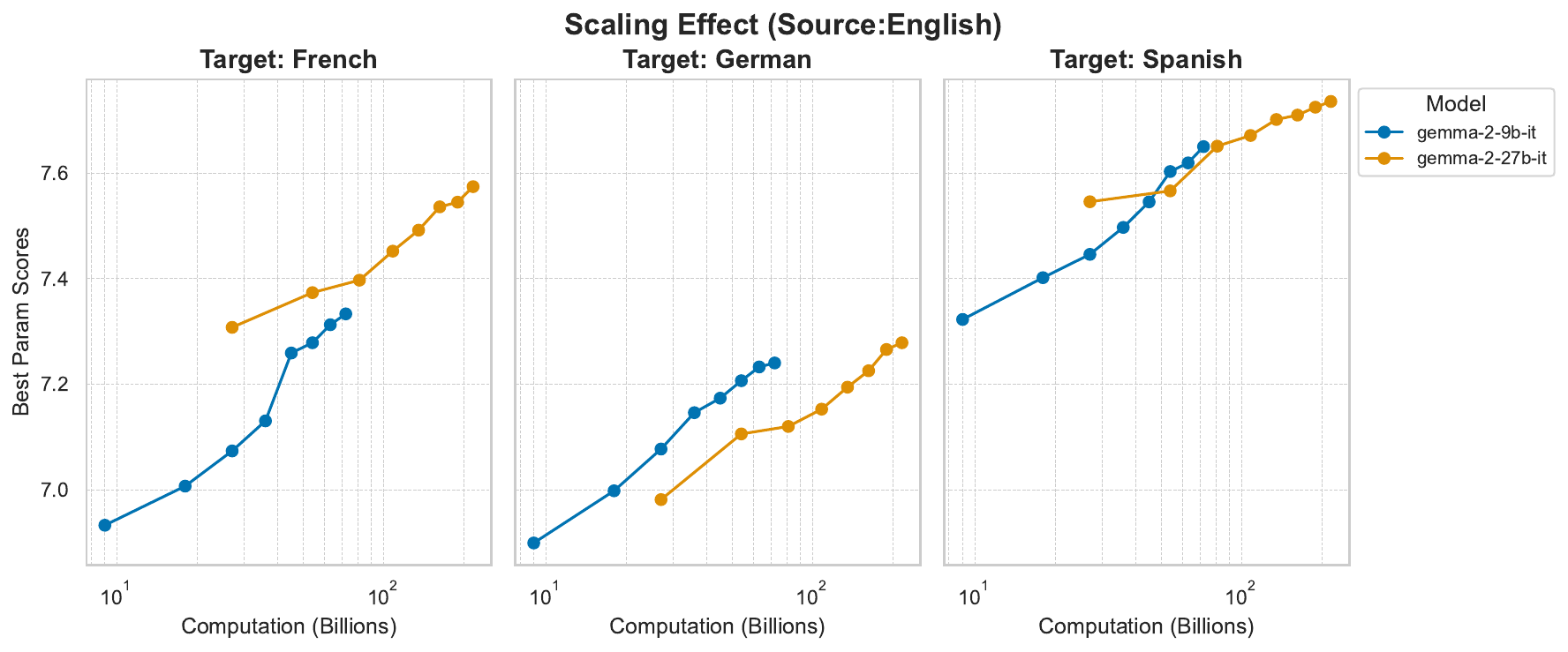}
    \caption{Cycle consistency score with respect to model size and inference computation.}
    \label{fig:scaling_effect}
\end{figure}

To evaluate the effectiveness of our cycle consistency framework for machine translation, we conducted experiments using range of available open-sourced LLMs with varying parameter sizes, and a carefully curated dataset of short paragraphs. The primary objectives were to assess how model scaling influences cycle consistency and to determine the suitability of validity of this metric.

The models used in our experiments include \textbf{Gemma-2} (9B and 27B parameters) [\cite{team2024gemma}], and \textbf{Qwen-2.5} (0.5B, 1.5B, 3B, 7B and 14B parameters) [\cite{yang2024qwen2}]. The transformer architecture, pretraining data and optimization pipeline between Gemma and Qwen should have some shared designs and some distinct features of themselves, leads to strength and weakness of each model series, so that we can examine our points more thoroughly. 

For each model, we set temperature of LLM sampling to be within $0.0, 0.15, 0.3, 0.45, 0.6,...$, and for the algorithm that generate $N$ candidates in forward translation, we temperature of of the $i$-th candidates to be $i * 0.15$,  so that for larger inference budge, we try to generate more diverse results for a better optimal candidate. For each forward translation, we use the same backward translation. For each complete procedure of translation, we calculate the computation consumed as the number of parameters in LLM, times number of forward translation, which coarsely reflect the overall computation. For a procedure of $N$ forward translation, each of them can be completed in parallel, as long as the LLM server scales.

We use ROUGE package for evaluating consistency. The package implements ROUGE-1, ROUGE-2, and ROUGE-L scores, each focuses on $1$-gram, $2$-gram and longest common subsequences. Each score include precision (\( p \)), recall (\( r \)), and F1-score (\( f \)) components. We aggregate the individual ROUGE scores by summing the precision, recall, and F1-scores across ROUGE-1, ROUGE-2, and ROUGE-L, leading to $[0,9]$ consistency score range, and $9.0$ being a perfect translation and a perfect LLM. For source languages such as Chinese or Japanese, which do not use spaces to separate words, we employ the \textit{jieba} library for tokenization. This step is crucial to accurately segment the text into meaningful tokens, ensuring that N-gram extraction is performed correctly.



\begin{lstlisting}[ label={Consistency Score Calculation}]{python}
    rouge = Rouge()
    scores = rouge.get_scores(original_text, cycle_result)
    score_sum = 0.0
    for score_k1 in ['rouge-1', 'rouge-2', 'rouge-l']:
        for score_k2 in ['r', 'p', 'f']:
            score_sum += scores[0][score_k1][score_k2]
\end{lstlisting}





\begin{figure}[h]
    \centering
    \includegraphics[width=\linewidth]{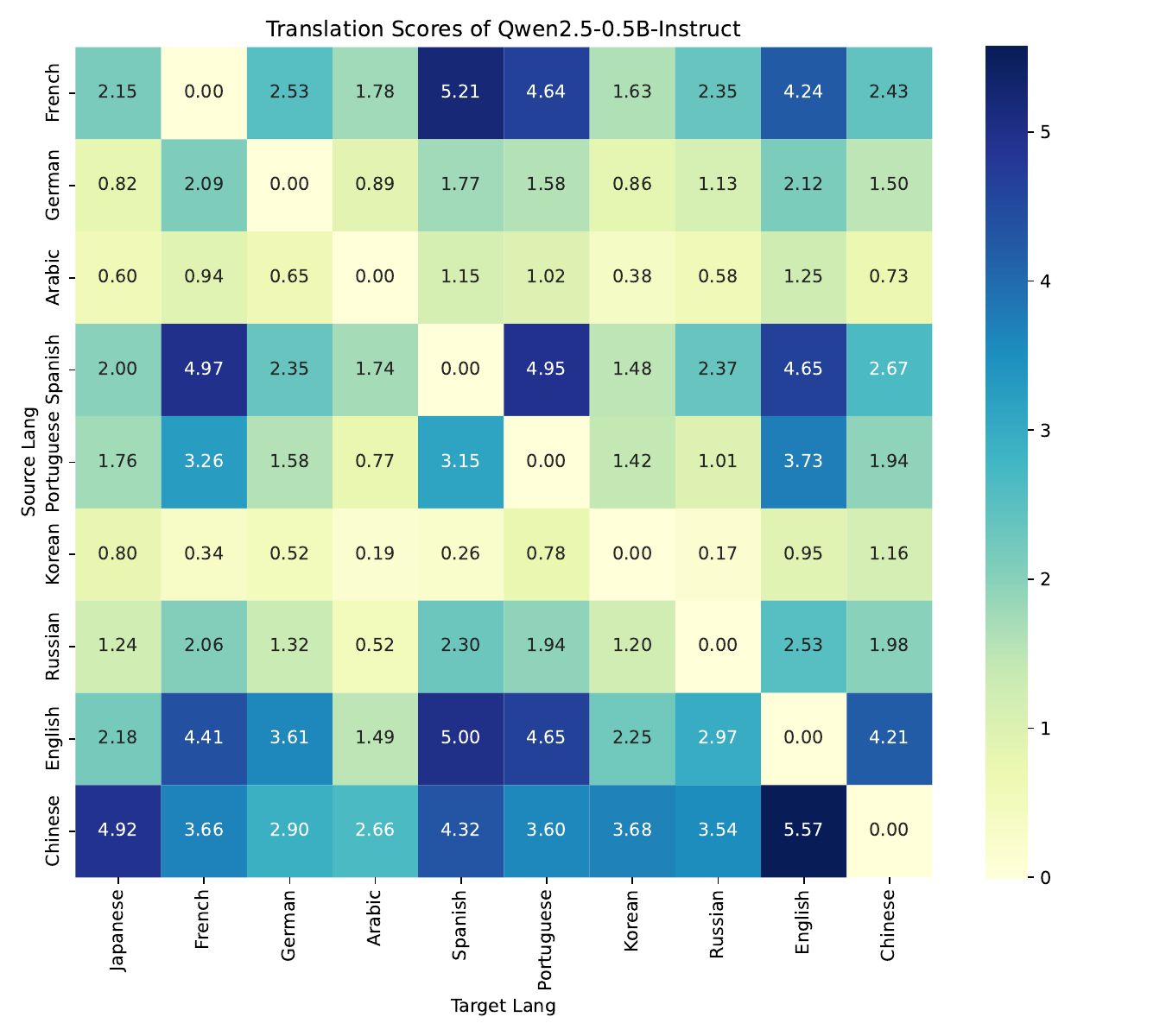}
    \vspace{-2em}
    \caption{Evaluation of any-to-any translation with an extremely small LLM.}
    \label{fig:any2any2}
\end{figure}

Our evaluation dataset consists of 100 short paragraphs written by GPT-4 on topics in diverse domains. Each paragraph is approximately 100 to 200 words long, ensuring a balance between complexity and manageability. The dataset was chosen to represent a wide range of topics and vocabulary, and they are unlikely to be included in the pretraining datasets, preventing overfitting. Some topics include: blockchain technology,  2024 U.S. Presidential Election, quantum computing, climate change affected global weather patterns, developments in artificial intelligence, latest advancements in renewable energy, impact of streaming services on the entertainment industry, etc.


\begin{figure}[h]
    \centering
    \includegraphics[width=\linewidth]{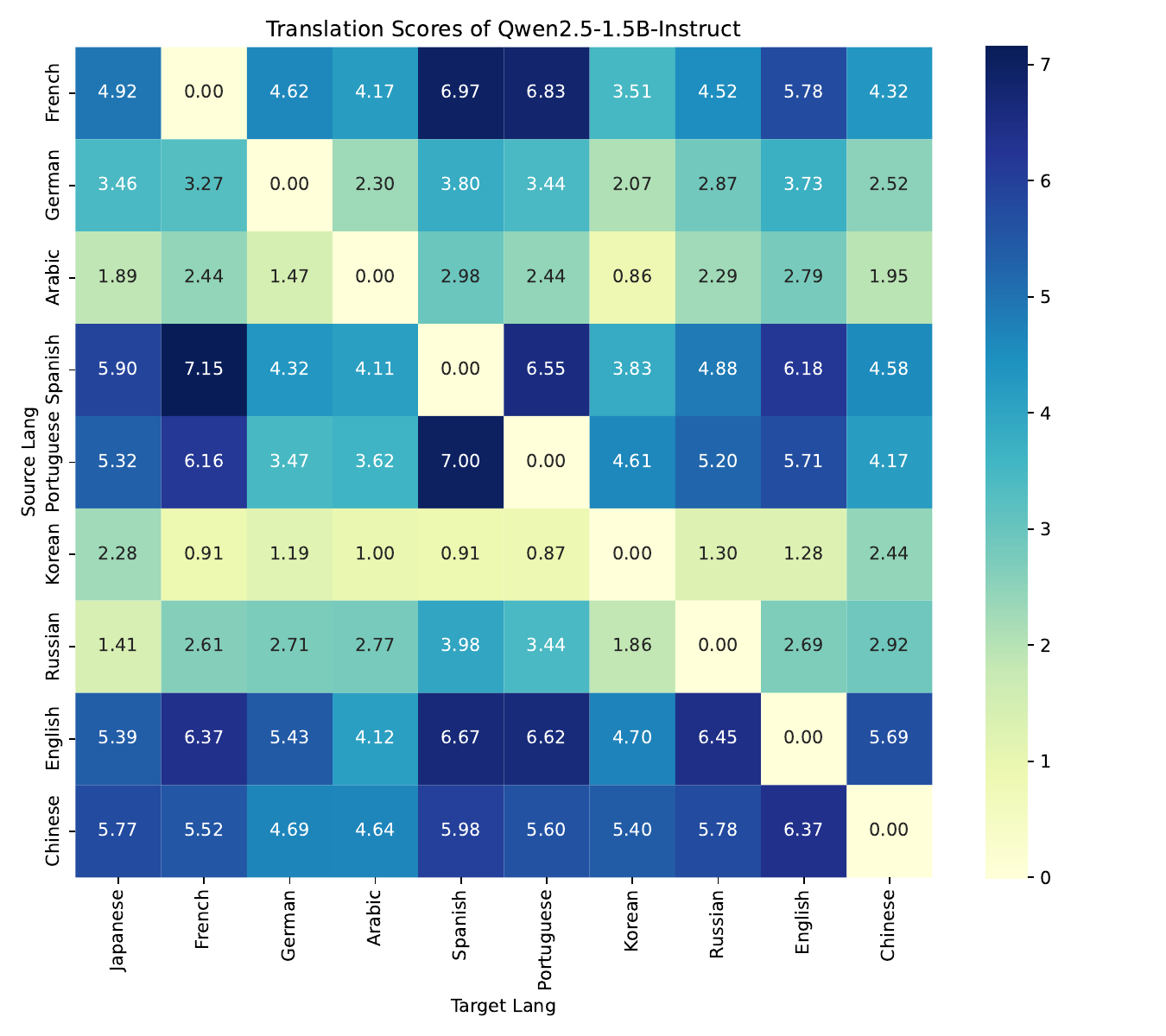}
    \vspace{-2em}
    \caption{Evaluation of any-to-any translation with a small LLM.}
    \label{fig:any2any3}
\end{figure}

\begin{figure}[h]
    \centering
    \includegraphics[width=\linewidth]{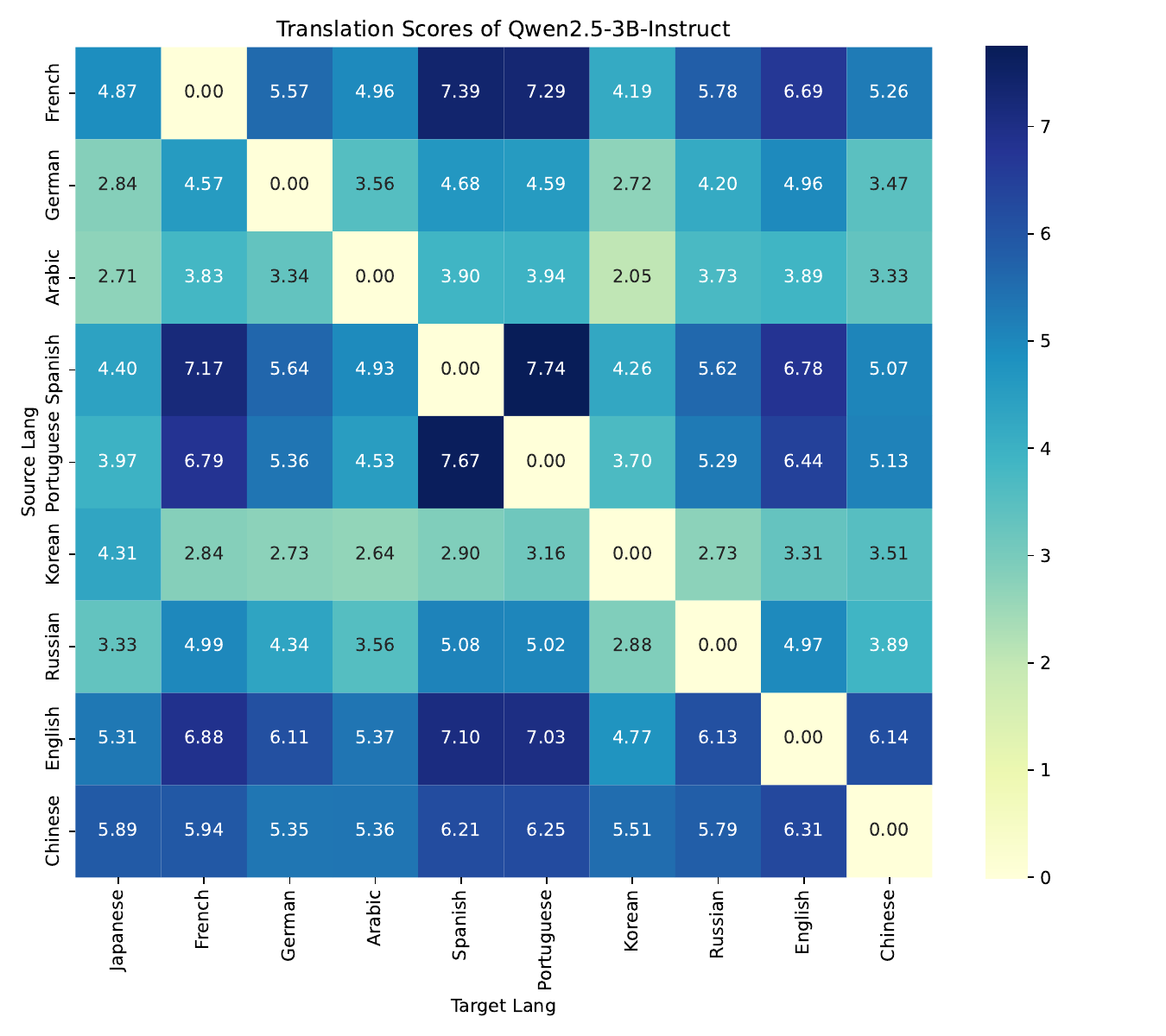}
    \vspace{-2em}
    \caption{Evaluation of any-to-any translation with a small LLM.}
    \label{fig:any2any4}
\end{figure}

\begin{figure}[h]
    \centering
    \includegraphics[width=\linewidth]{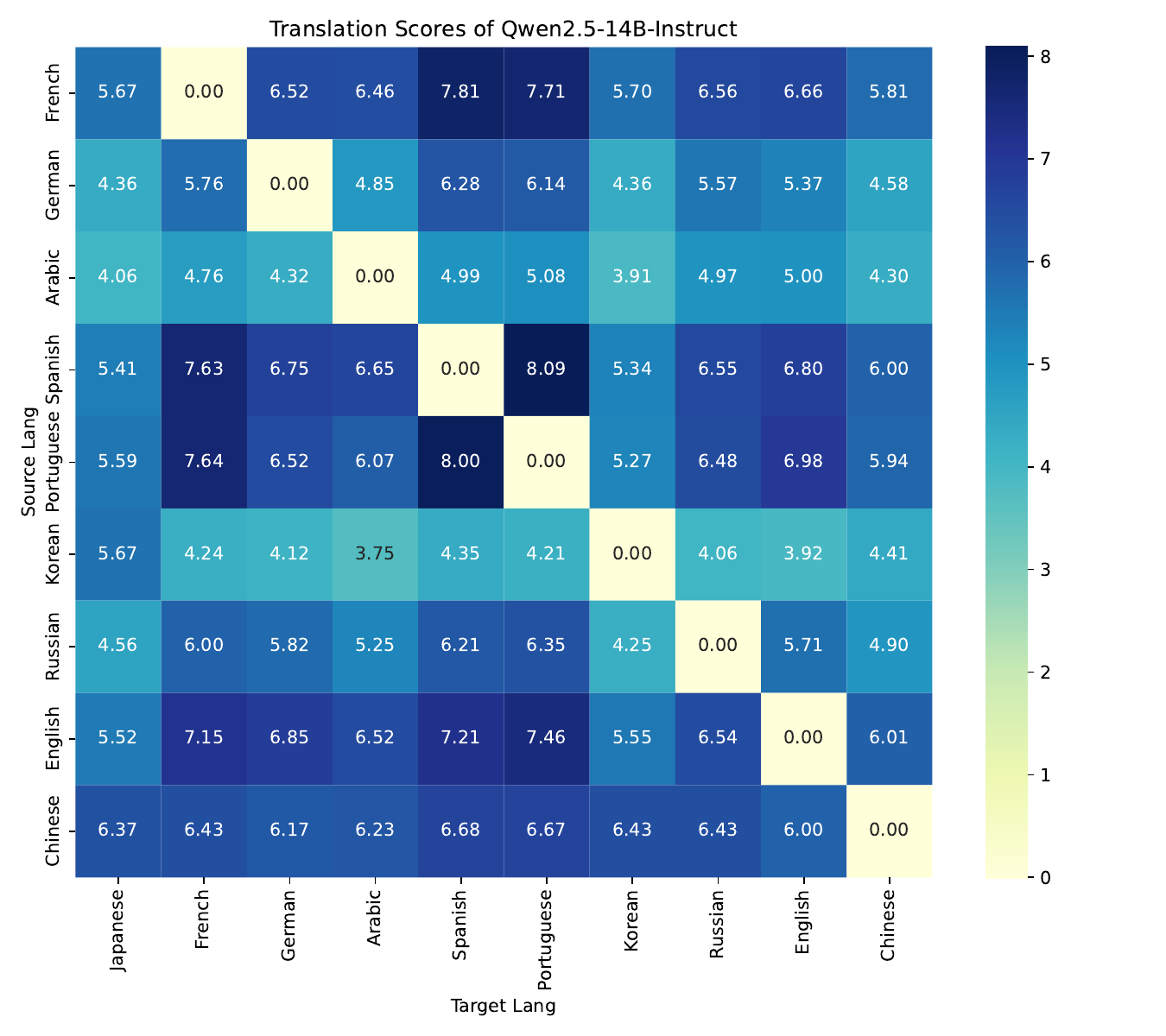}
    \vspace{-2em}
    \caption{Evaluation of any-to-any translation with a medium-sized LLM.}
    \label{fig:any2any5}
\end{figure}

The results (Figure \ref{fig:scaling_effect}) indicate a positive correlation between model size and cycle consistency. Larger models consistently achieve higher consistency. This trend aligns with the scaling laws of LLMs [\cite{kaplan2020scaling}], suggesting that increased model capacity enhances the ability to produce coherent and consistent translations. This  confirms that the cycle consistency is a meaningful metric for the translation quality, and the translation capability of LLM. If not, the cycle consistency score won't sharply increase as more forward translation being tried, nor a larger LLM being used. This can also be confirmed by Qwen model series as well in figures \ref{fig:any2any2} \ref{fig:any2any3}, \ref{fig:any2any4}, and \ref{fig:any2any5}.

In Figure \ref{fig:scaling_effect}, we visualize three representative patterns of scaling effects:

\begin{itemize}
    \item \textbf{Pattern 1:}  Using a small LLM repeatedly is useful but not the optimal choice if there is computational budget for running a larger LLM. Example: translation to French
    \item \textbf{Pattern 2:} Repeating a small LLM is effective and can outperform a larger LLM under the same computational budget. Example: translation to German
    \item \textbf{Pattern 3:} Repeating a small LLM is beneficial, and we could switch to a larger LLM if the marginal gain justifies the computational cost. Example: translation to Spanish.
\end{itemize}

The application of cycle consistency provides an easy way to evaluate LLM translations without paired sentences as labels. We ran benchmarks on translations from any language to any language, with extensive experiments using Qwen LLMs. In Figures \ref{fig:any2any3}, \ref{fig:any2any4}, and \ref{fig:any2any5}, we see that the translation between certain languages, such as Chinese and English, which have enough corpora during LLM pretraining, achieves good result even with 0.5B parameters, and the gain becomes marginal while increasing model size. While the LLM doesn't fit certain languages very well, we are still able to observe a interesting result: e.g. the translation between Spanish and Portuguese appears to be the most accurate, both forward and backward, in all sizes of Qwen LLMs. This arises from the fact that these two languages are quite similar to each other, both being part of the Romance language family that evolved from Latin. This shared origin results in numerous linguistic similarities, including vocabulary, grammar structures, and syntax. 

\section{Related Work}
Before the LLM era, similar ideas of cycle consistency were used for NLP tasks such as unsupervised machine translation, text style transfer, and dialogue systems. Early research used forward and backward translation to train models without parallel corpora [\cite{artetxe2018unsupervised}]. In text style transfer, it maintains the semantic content while altering stylistic attributes, ensuring that the transformed text remains faithful to the original meaning [\cite{shen2017style}]. Additionally, in dialogue systems, cycle consistency helps maintain conversational coherence by ensuring that responses can be accurately reverted to the original dialogue context [\cite{kim2018cycle}].

In speech processing, cycle consistency plays a crucial role in tasks like voice conversion and speech synthesis. Models leveraging cycle consistency can transform a source speaker's voice to a target speaker's voice without parallel data, ensuring that the converted voice can be reverted back accurately [\cite{lee2018unsupervised}]. This bidirectional consistency maintains the linguistic content while altering vocal characteristics, resulting in natural-sounding conversions. Moreover, in speech enhancement, cycle consistency helps in denoising by ensuring that enhanced speech can revert to its noisy form, thereby preserving speech quality and intelligibility [\cite{liu2022pre}]. These applications highlight the effectiveness of cycle consistency in maintaining speech integrity during transformations.

In computer vision, cycle consistency has been fundamental in various tasks by ensuring transformations retain essential information. In image-to-image translation, CycleGAN [\cite{zhu2017unpaired}] enables converting images between domains without paired examples, such as turning horses into zebras while maintaining structural integrity. For optical flow estimation, cycle consistency ensures that motion vectors from frame \( A \) to frame \( B \) can be accurately reversed, enhancing flow accuracy [\cite{baker2009database}]. In depth estimation, enforcing cycle consistency helps maintain geometric coherence, leading to more reliable depth maps [\cite{godard2017unsupervised}]. Additionally, in 3D object reconstruction, cycle consistency ensures that 3D models can be accurately projected back to their original 2D images, preserving structural details [\cite{schwarz2018group}]. These applications demonstrate how cycle consistency enhances the robustness and quality of computer vision models across diverse tasks.

\section{Conclusion}

This paper presents a self-reflective capabilities framework for machine translation that leverages large language models. 
Our findings demonstrate that larger models and more repetition both exhibit significant improvements in cycle consistency, highlighting the benefits of model size-scaling and inference computation-scaling in MT tasks. This should have  broader applications to reinforcement learning from human feedback (RLHF) and chain-of-thought reasoning (CoT). This approach offers a scalable and efficient alternative to traditional MT evaluation methods, paving the way for more intelligent and autonomous translation systems.



\appendix
\small{
\bibliography{ztx}
\bibliographystyle{ACM-Reference-Format}
}

\end{document}